\documentclass[11pt]{article}
\usepackage{graphicx}
\usepackage{amssymb}
\usepackage{amscd}
\usepackage{mathrsfs}
\usepackage{longtable,lscape}
\usepackage{amsthm}
\usepackage{amsfonts}
\usepackage{amsmath}
\usepackage{bbm}
\usepackage{float}
\usepackage{url}

\newcommand{\captionfonts}{\footnotesize}
\makeatletter  
\long\def\@makecaption#1#2{%
  \vskip\abovecaptionskip
  \sbox\@tempboxa{{\captionfonts #1: #2}}%
  \ifdim \wd\@tempboxa >\hsize
    {\captionfonts #1: #2\par}
  \else
    \hbox to\hsize{\hfil\box\@tempboxa\hfil}%
  \fi
  \vskip\belowcaptionskip}
\makeatother 
\begin{document}
\title{Quantum cognition beyond Hilbert space II: Applications}
\author{Diederik Aerts$^1$, Lyneth Beltran$^1$, Massimiliano Sassoli de Bianchi$^{2}$, Sandro Sozzo$^{3}$ and Tomas Veloz$^{1}$ \vspace{0.5 cm} \\ 
        \normalsize\itshape
        $^1$ Center Leo Apostel for Interdisciplinary Studies, 
         Brussels Free University \\ 
        \normalsize\itshape
         Krijgskundestraat 33, 1160 Brussels, Belgium \\
        \normalsize
        E-Mails: \url{diraerts@vub.ac.be,lyneth.benedictinelawcenter@gmail.com}
          \vspace{0.5 cm} \\ 
        \normalsize\itshape
        $^2$ Laboratorio di Autoricerca di Base \\
        \normalsize\itshape
        6914 Lugano, Switzerland
        \\
        \normalsize
        E-Mail: \url{autoricerca@gmail.com}
          \vspace{0.5 cm} \\ 
        \normalsize\itshape
        $^3$ School of Management and IQSCS, University of Leicester \\ 
        \normalsize\itshape
         University Road, LE1 7RH Leicester, United Kingdom \\
        \normalsize
        E-Mail: \url{ss831@le.ac.uk} 
       	\\
              }
\date{}
\maketitle
\begin{abstract}
\noindent
The research on human cognition has recently benefited from the use of the mathematical formalism of quantum theory in Hilbert space. However, cognitive situations exist which indicate that the Hilbert space structure, and the associated Born rule, would be insufficient to provide a satisfactory modeling of the collected data, so that one needs to go beyond Hilbert space. In Part I of this paper we follow this direction and  present a general tension-reduction (GTR) model, in the ambit of an operational and realistic framework for human cognition \cite{AertsSassolideBianchiSozzoVeloz2016}. In this Part II we apply this non-Hilbertian quantum-like model to faithfully reproduce the probabilities of the `Clinton/Gore' and `Rose/Jackson' experiments on question order effects. We also explain why the GTR-model is needed if one wants to deal, in a fully consistent way, with response replicability and unpacking effects. 
\end{abstract}

\section{Introduction\label{intro}}
The possibility of modeling data from human cognition by using the probabilistic formalism of quantum theory in Hilbert space has been suggested by one of us, and other authors, already two decades ago (see \cite{Khrennikov2010,bb2012,haven2013,Wendt2015} and the references therein). From the beginning it was emphasized that: ``in the same way as in geometry (where, starting with Lobachevsky, Gauss, Riemann, ..., various non-Euclidean geometries were developed and widely applied, e.g., in relativity theory), in probability theory various non-Kolmogorovian models may be developed to serve applications. The QM probabilistic model was one of the first non-Kolmogorovian models that had important applications. 
Thus one may expect development of other types of probabilistic models which would be neither Kolmogorovian nor quantum.'' \cite{Khrennikov2010}

In that respect, it was recently demonstrated that important cognitive situations exist which require to go beyond 
the quantum representation in Hilbert space, using more general probabilistic structures \cite{AertsSassolideBianchi2014a,AertsSassolideBianchi2015a,AertsSassolideBianchi2015b,asdb2015f,asdb2015a,AertsSassolideBianchiSozzo2015,Boyer-KassemEtal2015,Boyer-KassemEtal2015b}. However, going beyond quantum does not mean to renounce an operational and realistic description of the cognitive entities, as prescribed by the Brussels' approach to cognition \cite{AertsSassolideBianchiSozzo2015}. It simply means adopting an extended (or completed) version of quantum theory, where different probability assignments, in a addition to the Born rule, can be described and coherently integrated in the formalism, and where state spaces different than the Hilbert space can also be envisaged. 

This is what we have recently done in the derivation of the so-called `general tension-reduction (GTR) model' and the associated `extended Bloch representation (EBR)' of quantum theory,  which we describe in Part I of this paper in the framework of sequential dichotomic measurements  \cite{AertsSassolideBianchiSozzoVeloz2016}. In this Part II, after recalling the basic elements of the model (Sec. \ref{modeling-sequential}), we apply  it to obtain an exact (and not only approximate) representation of the opinion poll data collected by Moore \cite{Moore2002} and exhibiting question order effects (Sec.~\ref{modeling-Moore}). We then show in Sec.~\ref{replicability} how question order effects and response replicability can be modeled together within the GTR-model, which is not the case in the Hilbert space quantum modeling 
\cite{KhrennikovEtal2014,KhrennikovBasieva2015}. Finally, we observe that another cognitive effect, the `unpacking effect', 
also requires a non-Kolmogorovian probability framework, like the one provided by the GTR-model, when unpacking effects are interpreted in terms of the relationship between measurements and sub-measurements.

\section{Modeling data on sequential measurements\label{modeling-sequential}}

The GTR-model describes measurements by taking into account the presence of irreducible and uncontrollable fluctuations in the experimental context, giving rise to a weighted symmetry breaking process that selects, in a way that cannot in general be known in advance, one of the available outcomes \cite{AertsSassolideBianchi2015a,AertsSassolideBianchi2015b,asdb2015f}. 
When the state space is a Hilbert space, the model uses a `generalized Bloch representation', which in the special case of measurements with two outcomes, or dichotomic, reduces to the well-known 3-dimensional Bloch sphere. Also, when the fluctuations are assumed to be uniform, the model provides the same predictions of the Born rule of quantum probability, of which it is therefore a natural generalization \cite{AertsSassolideBianchi2014a}. 

In \cite{AertsSassolideBianchiSozzoVeloz2016} we use the GTR-model and its EBR implementation to represent cognitive entities and sequential dichotomic measurements on these entities. More precisely, the two-dimensional Hilbert vector $|\psi\rangle$ describing the initial state of a cognitive entity is represented, in the EBR representation, by a 3-dimensional real vector ${\bf x}_\psi$, and if a dichotomic measurement $A$ has the outcomes $A_y$ and $A_n$, the associated  probabilities $p_\psi(A_y)$ and $p_\psi(A_n)$ are described by the two integrals
\begin{equation}
p_\psi(A_y)=\int_{-1}^{\cos\theta_A}\rho_A(x|\psi) dx\quad\quad p_\psi(A_n)=\int_{\cos\theta_A}^{1}\rho_A(x|\psi) dx
\label{A-integrals}
\end{equation}
where $\rho_A(x|\psi)$ is a probability distribution on the line segment $[-1,1]$, associated with the measurement $A$, which in general may also depend on the initial state, and $\theta_A$ is the angle between ${\bf x}_\psi$ and the vector ${\bf a}_y$ representing the entity's state associated with the outcome $A_y$, on the Bloch sphere (${\bf x}_\psi\cdot {\bf a}_y=\cos\theta_A$); for the interpretation of (\ref{A-integrals}) see for instance \cite{AertsSassolideBianchiSozzoVeloz2016}. 

The Born rule is clearly recovered when,  for every initial state, 
$\rho_A(x|\psi)={1\over 2}$, so that more general probabilistic rules can easily be described by adopting non-uniform probability distributions $\rho_A(x|\psi)$. To have an explicitly solvable model, we assume in the following that $\rho_A(x|\psi)$ does not depend on the initial state and that it is `locally uniform', i.e. only characterized by two parameters $\epsilon_A \in [0,1]$ and $d_A\in [-1+\epsilon_A, 1-\epsilon_A]$, such that: $\rho_A(x) = 0$ if $x\in [-1, d_A- \epsilon_A)\cup (d_A+ \epsilon_A,1]$, and $\rho_A(x) = 1/2\epsilon_A$ if $x \in [d_A- \epsilon_A,d_A+ \epsilon_A]$. To obtain compact expressions, we also assume that $\cos\theta_A \in [d_A- \epsilon_A,d_A+ \epsilon_A]$. 

If we describe in a similar way a second dichotomic measurement $B$, then in addition to the three parameters $\epsilon_A$, $d_A$ and $\theta_A$, characterizing $A$, we have three more  parameters $\epsilon_B$, $d_B$ and $\theta_B$, characterizing $B$, and a supplementary parameter $\theta$, defined by $\cos\theta = {\bf a}_y\cdot {\bf b}_y$, characterizing the relative orientation of the two measurements within the Bloch sphere. In the following we also assume that $\cos\theta\in [d_A- \epsilon_A,d_A+ \epsilon_A]$ and $\cos\theta \in [d_B- \epsilon_B,d_B+ \epsilon_B]$. Then, if we
perform in sequence the measurement $A$ followed by the measurement $B$ (which we denote $AB$), the sequential measurement has the 4 outcomes $A_iB_j$, $i,j\in\{y, n\}$, and the associated probabilities are given by the products $p_\psi(A_iB_j)=p_\psi(A_i)p_{A_i}(B_j)$. Then, performing the integrals (\ref{A-integrals}), one obtains
\begin{eqnarray}
&&p_\psi(A_yB_y)={1\over 4}(1+{\cos\theta -d_B\over \epsilon_B})(1+{\cos\theta_A -d_A\over \epsilon_A}) \nonumber\\
&&p_\psi(A_yB_n)={1\over 4} (1-{\cos\theta -d_B\over \epsilon_B})(1+{\cos\theta_A -d_A\over \epsilon_A})\nonumber\\
&&p_\psi(A_nB_n)={1\over 4} (1+{\cos\theta +d_B\over \epsilon_B})(1-{\cos\theta_A -d_A\over \epsilon_A})\nonumber\\
&&p_\psi(A_nB_y)={1\over 4} (1-{\cos\theta +d_B\over \epsilon_B})(1-{\cos\theta_A -d_A\over \epsilon_A}).
\label{explicitsolutionAB}
\end{eqnarray}
Similarly, for the reversed order sequential measurement $BA$, where we first perform $B$ and then $A$, we have
\begin{eqnarray}
&&p_\psi(B_yA_y)={1\over 4}(1+{\cos\theta -d_A\over \epsilon_A})(1+{\cos\theta_B -d_B\over \epsilon_B}) \nonumber\\
&&p_\psi(B_yA_n)={1\over 4} (1-{\cos\theta -d_A\over \epsilon_A})(1+{\cos\theta_B -d_B\over \epsilon_B})\nonumber\\
&&p_\psi(B_nA_n)={1\over 4} (1+{\cos\theta +d_A\over \epsilon_A})(1-{\cos\theta_B -d_B\over \epsilon_B})\nonumber\\
&&p_\psi(B_nA_y)={1\over 4} (1-{\cos\theta +d_A\over \epsilon_A})(1-{\cos\theta_B -d_B\over \epsilon_B}).
\label{explicitsolutionBA}
\end{eqnarray}

The above system of equations is underdetermined, in the sense that the 8 outcome probabilities can determine all the parameters but one. This means that we are free to choose one of the parameters, for instance $\epsilon_A$, and by doing so all the others will be automatically fixed. We observe that since we must have $\epsilon_A(1+{d_A\over \epsilon_A})\leq 1$, i.e. $\epsilon_A\leq 1/(1+{d_A\over \epsilon_A})$, this means that if ${d_A\over \epsilon_A}$ is different from zero, it is not be possible to model the data by means of the standard quantum formalism (in a 2-dimensional Hilbert space), as the Born rule corresponds to the choice $d_A=0$ and $\epsilon_A =1$.

\section{Modeling Moore's data\label{modeling-Moore}}
We now use (\ref{explicitsolutionAB})-(\ref{explicitsolutionBA}) to `exactly' model the data obtained in a Gallup poll conducted in 1997, as presented in a review of question order effects by Moore \cite{Moore2002}. More precisely, we consider the probabilities given by \cite{WangBusemeyer2013} (see also \cite{WangEtal2014}), where the participants who did not provided a `yes' or `no' answer have been excluded from the statistics. In one of the experiments, a thousand participants were subjected to an opinion poll, consisting of a pair of questions asked in a sequence. The first question, which we associate with  measurement $A$, is: ``Do you generally think Bill Clinton is honest and trustworthy?'' The second question, which we associate with measurement $B$, is: ``Do you generally think Al Gore is honest and trustworthy?'' Half of the participants were submitted to the two questions in the order $AB$ (first `Clinton' then `Gore') and the other half in the reversed order $BA$, and the collected response probabilities are:\footnote{Because of a rounding error, the probabilities given in \cite{WangBusemeyer2013} do not exactly sum to $1$, but to $0.9999$, in the $AB$ measurement, and to $1.0001$, in the $BA$ measurement. Since our solution requires them to exactly sum to $1$, we have corrected the value of $p(A_nB_n)$, from $0.2886$ to $0.2887$, and the value of $p(B_nA_n)$, from $0.2130$ to $0.2129$.} $p(A_yB_y)=0.4899$, $p(A_yB_n)= 0.0447$, $p(A_nB_y)=0.1767$, $p(A_nB_n)= 0.2887$, $p(B_yA_y)=0.5625$, $p(B_yA_n) = 0.1991$, $p(B_nA_y) = 0.0255$, $p(B_nA_n) = 0.2129$.

These probabilities show a significant question order effect. Inserting  them in (\ref{explicitsolutionAB})-(\ref{explicitsolutionBA}), one obtains, after some calculations, the following explicit values for the model's parameters (see \cite{asdb2015a} for a detailed analysis)
\begin{eqnarray}
&&{d_A\over \epsilon_A}=0.1545 \quad\quad {\cos\theta_A\over \epsilon_A}=0.2237 \quad\quad {\cos\theta \over \epsilon_A}=0.6316\nonumber\\
&&{d_B\over \epsilon_B}=-0.2961 \quad\quad {\cos\theta_B\over \epsilon_B}=0.2271 \quad\quad {\cos\theta \over \epsilon_B}=0.5367.
\end{eqnarray} \label{solutionsCG}
We immediately see that the solution does not admit a representation by means of the Born rule, considering that ${d_A\over \epsilon_A},{d_B\over \epsilon_B}\neq 0$. Furthermore, we see that we cannot have $\epsilon_A=\epsilon_B$ and $d_A=d_B$, i.e. the solution requires the two measurements to be characterized by different rules of probabilistic assignment ($\rho_A\neq\rho_B$). The structure of the probabilistic data is thus irreducibly non-Hilbertian. If we choose $\epsilon_A=1/2$, we obtain for the other parameters (writing them in approximate form, to facilitate their reading): $\epsilon_A=0.5$, $\epsilon_B\approx 0.59$, $d_A \approx 0.08$, $d_B\approx-0.17$,
$\cos\theta \approx 0.32$, $\cos\theta_A\approx 0.11$, $\cos\theta_B\approx 0.13$.

In another experiment reported by Moore, always performed on a thousand participants, the opinion poll consisted in a pair of questions about the baseball players Pete Rose and Shoeless Joe Jackson. More precisely, the question $A$ was: ``Do you think Rose should or should not be eligible for admission to the Hall of Fame?''. Similarly, the question $B$ was: ``Do you think Jackson should or should not be eligible for admission to the Hall of Fame?''. The collected response probabilities are (also in this case we use the probability data given in \cite{WangBusemeyer2013,WangEtal2014}): $p(A_yB_y)=0.3379$, $p(A_yB_n)= 0.3241$, $p(A_nB_y) =0.0178$, $p(A_nB_n) = 0.3202$, $p(B_yA_y) =0.4156$, $p(B_yA_n) = 0.0671$, $p(B_nA_y) = 0.1234$, $p(B_nA_n) = 0.3939$, and the modeling now gives \cite{asdb2015a}
\begin{eqnarray} \label{solutionsRJ}
&&{d_A\over \epsilon_A}= -0.0995 \quad\quad {\cos\theta_A\over \epsilon_A}=0.2245 \quad\quad {\cos\theta \over \epsilon_A}=0.6224 \nonumber\\
&&{d_B\over \epsilon_B}=0.4369 \quad\quad {\cos\theta_B\over \epsilon_B}=0.4023 \quad\quad {\cos\theta \over \epsilon_B}=0.4578.
\end{eqnarray}

\begin{figure}[!ht]
\centering
\includegraphics[scale =.28]{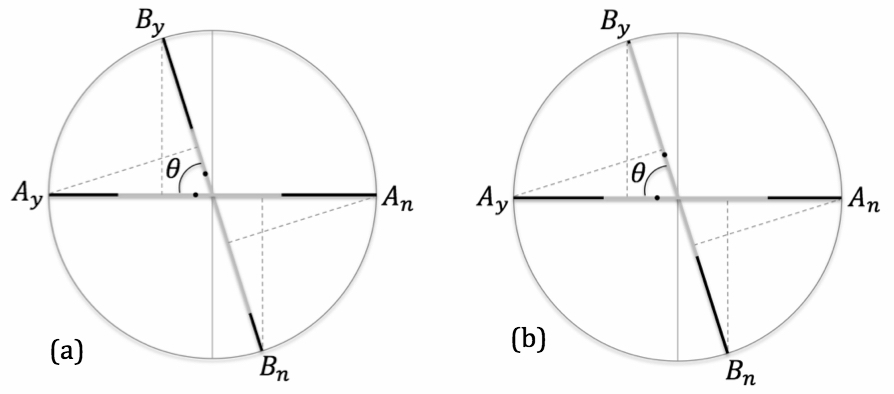}
\caption{The probability distributions describing (a) the data of the Clinton/Gore and (b) of the Rose/Jackson experiments in the GTR-model.
\label{Clinto-Gore-Rose-Jackson}}
\end{figure}

Again, we observe that these values are irreducibly non-Hilbertian. For $\epsilon_A=1/2$, we obtain: $\epsilon_A=0.5$, $\epsilon_B\approx 0.68$, $d_A \approx -0.05$, $d_B\approx 0.30$, $\cos\theta \approx 0.31$, $\cos\theta_A\approx 0.11$, $\cos\theta_B \approx 0.27$. In Fig.~\ref{Clinto-Gore-Rose-Jackson} the two solutions (\ref{solutionsCG}) and (\ref{solutionsRJ})
are graphically represented. The black dots denote the values of $\cos\theta_A$ and $\cos\theta_B$, and the black regions are those where the probability distributions is zero (corresponding to the unbreakable elastic regions \cite{AertsSassolideBianchiSozzoVeloz2016}).

What strikes the eye is that the solutions (\ref{solutionsCG}) and (\ref{solutionsRJ})  are structurally similar, despite of the fact that only the former (almost) obey the so-called QQ equality \cite{WangBusemeyer2013,WangEtal2014}:
$q \equiv p_\psi(A_yB_y)-p_\psi(B_yA_y)+p_\psi(A_nB_n) -p_\psi(B_nA_n)=0$.  This is because the latter is insufficient to fully characterize a Hilbertian structure and that both solutions are actually intrinsically non-Hilbertian \cite{asdb2015a}. 

It is worth mentioning that the QQ equality follows from the simple operatorial identity \cite{asdb2015a,AertsSassolideBianchiSozzo2015}: $Q\equiv P_y^AP_y^BP_y^A - P_y^BP_y^AP_y^B + P_n^AP_n^BP_n^A - P_n^B P_n^AP_n^B=0$, as is clear that $p_\psi(A_iB_j)= \langle \psi|P_i^AP_j^BP_i^A|\psi\rangle$ and $p_\psi(B_jA_i)= \langle \psi|P_j^BP_i^AP_j^B|\psi\rangle$, where $P_i^A$ and  $P_j^B$ are the projection operators onto the states associated with the outcomes 
$A_i$ and $B_j$, respectively, $i,j\in\{y,n\}$. Now, it has been pointed out that the Clinton/Gore data are different from the Rose/Jackson as for the latter participants also received some sequential background information before answering the two questions, and this would explain why, contrary to the Clinton/Gore data, they disobey the QQ equality. Indeed, if this supply of information is modeled by using two unitary operators $U$ (for the information given before $A$) and $V$ (for that given before $B$), we now have to write $p_\psi(A_iB_j)=\langle \psi|U^\dagger P_i^A V^\dagger P_j^BVP_i^AU|\psi\rangle$, and similarly for $p_\psi(B_jA_i)$. Thus, the relevant operator becomes:
\begin{eqnarray}
Q'&=&U^\dagger P_y^A{P'}_y^BP_y^AU - V^\dagger P_y^B{P'}_y^AP_y^BV + U^\dagger P_n^A{P'}_n^BP_n^AU - V^\dagger P_n^B {P'}_n^AP_n^BV\nonumber\\
&=& [{P'}_y^B - U^\dagger {P'}_y^BU] +[V^\dagger {P'}_y^AV -{P'}_y^A] + [U^\dagger {P'}_y^BU {P'}_y^A - {P'}_y^BV^\dagger {P'}_y^AV]\nonumber\\
&&\quad +\, [{P'}_y^AU^\dagger {P'}_y^BU - V^\dagger {P'}_y^AV {P'}_y^B]
\end{eqnarray}
where we have defined ${P'}_i^A\equiv U^\dagger P_i^AU$, ${P'}_j^B\equiv V^\dagger P_j^BV$, $i,j\in\{y,n\}$. Since the average $\langle \psi|Q'|\psi\rangle$ can now in principle take any value within the interval $[-1,1]$ (unless $U=V=\mathbb{I}$), this could explain why the QQ equality is disobeyed in the Rose/Jackson measurement. 

The above argument, however, is weakened by the observation that there are other quantum equalities that are strongly disobeyed both by the Clinton/Gore and Rose/Jackson data, like for instance, in the situation of non-degenerate measurements \cite{asdb2015a,AertsSassolideBianchiSozzo2015}: $q'\equiv p_\psi(A_yB_n)p_\psi(A_nB_n) - p_\psi(A_nB_y)p_\psi(A_yB_y)=0$, which must be obeyed also when participants receive some background information. Indeed, we have in this case $p_\psi(A_iB_j) = |\langle A_i|U|\psi\rangle|^2 |\langle B_j|V|A_i\rangle |^2$, where $P_i^A=|A_i\rangle\langle A_i|$ and $P_j^B=|B_j\rangle\langle B_j|$, $i,j\in\{y,n\}$, so that we can write:
\begin{eqnarray}
&&q'= |\langle A_y|U|\psi\rangle|^2 |\langle A_n|U|\psi\rangle|^2 \times\nonumber\\
&&\quad \quad \quad \times \left[ |\langle B_n|V|A_y\rangle |^2 |\langle B_n|V|A_n\rangle |^2 - |\langle B_y|V|A_n\rangle |^2 |\langle B_y|V|A_y\rangle |^2\right]
\label{q-prime2}
\end{eqnarray}
Using $|\langle B_y|V|A_n\rangle |^2 = 1- |\langle B_n|V|A_n\rangle |^2$ and $|\langle B_y|V|A_y\rangle |^2=1-|\langle B_n|V|A_y\rangle |^2$, it is easy to check  that the terms in the above bracket cancel, so that $q'=0$. Thus, we have a pure quantum equality which must be obeyed also when some information is sequentially provided to the participants. However, it is strongly violated by the experimental data \cite{asdb2015a}.

\section{Response replicability\label{replicability}}
As emphasized in \cite{KhrennikovEtal2014}, the standard quantum formalism is unable to jointly model question order effects and response replicability. The reason is simple to understand: response replicability, the situation where a question, if asked a second time, receives the same answer, even if other questions have been answered in between, requires commuting observables to be modeled. Indeed, since we have the operatorial identity $P_n^BP_y^AP_n^B-P_y^AP_n^BP_y^A=(P_y^B-P_y^A)[P_y^B,P_y^A]$, it follows that the difference $p_\psi(B_nA_y)-p_\psi(A_yB_n)$ can generally be non-zero only if $[P_y^B,P_y^A]\neq 0$, i.e. the spectral families associated with the $A$ and $B$ measurements do not commute. So, not only an exact description of question order effects requires to go beyond-quantum, but the combination of the latter with 
response replicability also creates a contradiction, which persists even when measurements are represented by positive-operator valued measures \cite{KhrennikovEtal2014,KhrennikovBasieva2015}. 

The reason why the above contradiction cannot be eliminated is that in quantum theory an observable automatically determines, via the Born rule, the outcome probabilities. This means that, once the initial state is given, and the possible outcomes are also given, there is only one way to choose them: that prescribed by the Born rule. This means that if a specific participant would be able to interact with a cognitive entity by employing different ways of choosing, at least one of them has to be non-Bornian. In our opinion, such a situation precisely occurs when considering the effect of response replicability. Indeed, in this case there are at least two possible ways of choosing an outcome from the memory of the previous interaction. In the standard formalism there is no place to describe such a memory effect, hence the impossibility to model it in a consistent way, beyond the so-called `adjacent replicability' \cite{KhrennikovBasieva2015}, which is built-in in all first-kind measurements. 

On the other hand, in the richer structure of the GTR-model, changes in the way outcomes are selected can be easily modeled as changes in the measurements' probability distributions \cite{asdb2015a}. In other terms, the reason why `separated replicability' can be taken into account in the GTR-model, jointly with possible question order effects, is that it allows not only to describe how the action of contexts can produce state transitions, but also how state transitions can determine a change of future contexts, via a change of the associated probability distributions. This, as we said, would be impossible to describe within the standard formalism, as then the probability distributions must all be uniform. But we humans do not function like this: apart from exceptions, once we have formed an opinion we do not need to form it again, by definition of what an opinion (and consequently an opinion poll) is. 

\begin{figure}[!ht]
\centering
\includegraphics[scale =.28]{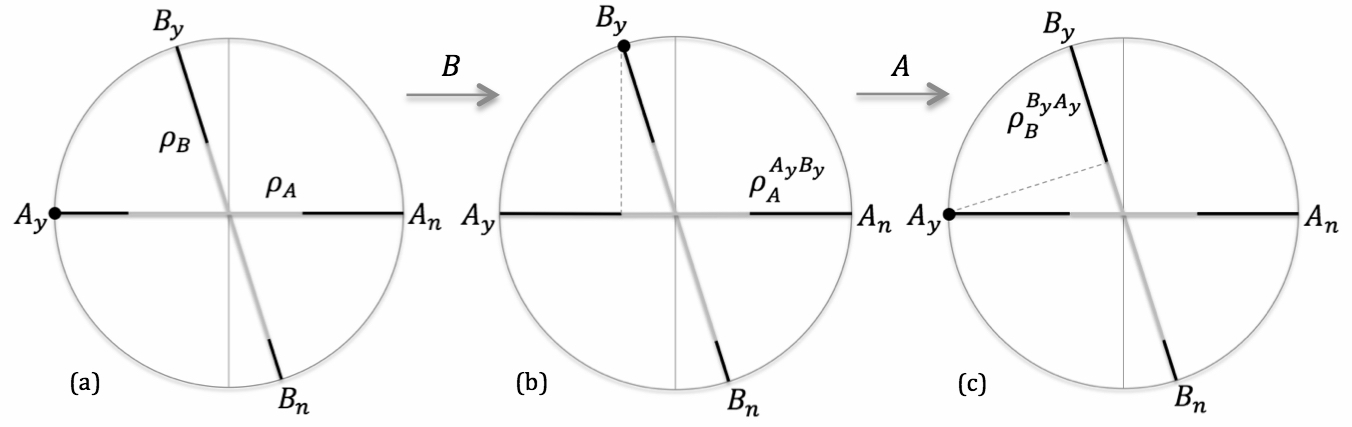}
\caption{The measurement sequence $BA$, in the GTR-model. 
\label{ABA}}
\end{figure}

To see how the above works, let us consider the sequence of three measurements $ABA$ on a cognitive entity (see \cite{asdb2015a} for a more general discussion). Let   $\rho_A$ be  the probability distribution describing the measurement $A$, and let us suppose that the outcome is $A_y$.  We do not need to associate any change of the probability distribution $\rho_A$ to this transition,  as measurements are already first kind measurements
in the GTR-model, as in quantum theory. Then, let us suppose that, when the measurement $B$ is performed, with the entity now in the state associated with outcome $A_y$, the outcome $B_y$ is obtained. Again,  we do not need to associate any change of the probability distribution $\rho_B$ to 
this second transition,  but we now have to update the probability distribution describing the measurement $A$, to guarantee that, if we repeat the latter, the outcome $A_y$  is  certain in advance. In other terms,  we now associate a probability distribution transition from $\rho_A$  to $\rho'_A$, able to ensure response replicability. And similarly, when the measurement $A$ is performed, giving $A_y$ with probability 1, there will be a probability distribution change from $\rho_B$ to $\rho'_B$, to ensure that a subsequent measurement $B$ will give $B_y$ with certainty, and from that point on subsequent $A$ or $B$ measurements can only deterministically reproduce the same outcomes, with no further changes of contexts. More precisely, the probability distributions $\rho'_A$ and $\rho'_B$  can be obtained by simple truncation and renormalization \cite{asdb2015a}:
\begin{equation}
\rho'_A(x) = {\rho_A\over \int_{-1}^{\cos\theta} \rho_Adx} \chi_{[-1,\cos\theta)}(x)\quad\quad \rho'_B(x) = {\rho_B\over \int_{-1}^{\cos\theta} \rho_Bdx} \chi_{[-1,\cos\theta)}(x)
\label{rhotransitions2}
\end{equation}
where $\chi_I(x)$ is the characteristic function of the interval $I$. Fig.~\ref{ABA} illustrates this `double transition process', where not only states but also probability distributions can change.  Figure (a) represents the situation following the first measurement $A$, the outcome being $A_y$. Figure (b) describes the subsequent measurement $B$, the outcome being $B_y$, also producing the  
transition from  $\rho_A$ to $\rho'_A$. Figure (c) describes the second measurement $A$, giving again outcome $A_y$, with certainty, which is also accompanied by the 
transition from $\rho_B$ to $\rho'_B$.

\section{Unpacking effects\label{unpacking}}
We analyze in this section the so-called `unpacking effects', usually modeled in the quantum formalism by assuming that the participants actually perform non-compatible sequential measurements, in a predetermined order \cite{Franco2008}. Our thesis is that, if we consider these effects in relation to the notion of sub-measurement, they point to an inadequacy of the quantum formalism in Hilbert space, as they describe situations that are incompatible with the quantum representation of degenerate measurements. 

Two kinds of unpacking are usually considered, `implicit' and `explicit'. The implicit unpacking is when a question is addressed in two different ways, a `packed way' and an `unpacked way'. More precisely, if $A$ and $B$ are two dichotomic measurements with outcomes $A_y$ and $A_n$, and $B_y$ and $B_n$, respectively, we can define a measurement $A'$, with outcomes $A'_y$ and $A'_n$, where $A'_n$ is the same as $A_n$, and $A'_y$ describes a possibility that is logically equivalent to $A_y$, expressed as an alternative over two mutually exclusive and exhaustive possibilities, defined by the outcomes of $B$. In other terms, $A'_y=(A_y\wedge B_y)\oplus (A_y\wedge B_n)$, where the symbol $\oplus$ denotes the logical exclusive conjunction.

An example adapted from a list of paradigmatic experiments performed by Rottenstreich and Tversky \cite{RottenstreichTversky1997} is the following. The measurement $A$ is the question: ``Is the winner of next US presidential election a non-Democrat?'', with outcome $A_y$ corresponding to the answer ``Yes, is a non-Democrat,'' and outcome $A_n$ corresponding to the answer ``No, is a Democrat.'' The measurement $B$ is the question: ``If the winner of next US presidential election is a non-Democrat, will be an Independent?'', with outcome $B_y$ corresponding to the answer ``Yes, an Independent,'' and outcome $B_n$ corresponding to the answer ``No, not an Independent.'' On the other hand, the implicitly unpacked measurement $A'$ is defined by the question: ``Is the winner of the next presidential election an Independent or Republican rather than a Democrat?'', with outcome $A'_y$ corresponding to the (unpacked) answer ``Yes, is an Independent or a Republican rather than a Democrat'' and outcome $A'_n$ to the answer ``No, is a Democrat,'' which is the same as $A_n$. 

Following Sec. 3 of \cite{AertsSassolideBianchiSozzoVeloz2016}, we denote by $p_S$ the initial state of the conceptual entity $S$ -- {\it  The winner of next US presidential election}, in our case. Moreover,   we denote by $p_{A_i}$ and $p_{A'_i}$ the final states of $S$  associated with the outcomes $A_i$ and $A_i'$, $i\in\{y,n\}$,
respectively. Then, we can write the corresponding probabilities as $p_S(A_i)=\mu(p_{A_i}, e_A,p_S)$ and  $p_S(A'_i)=\mu(p_{A'_i}, e_{A'},p_S)$, $i\in\{y,n\}$, where $e_A$ and $e_{A'}$ are the contexts associated with $A$ and $A'$, respectively, causing the transitions from the initial state $p_S$ to the observed outcome states $p_{A_i}$ and $p_{A'_i}$, respectively.  If $p_S(A'_y)$ is found to be sensibly different from $p_S(A_y)$, one says that there is an unpacking effect, i.e. an effect where logically equivalent descriptions of a same possibility can produce different probabilities, thus violating the so-called principle of `description invariance'. More precisely, one speaks of `superadditivity' if $p_S(A_y)>p_S(A'_y)$ and `subadditivity' if $p_S(A_y)<p_S(A'_y)$.

Let us also describe the situation corresponding to the `explicit unpacking effect'. In this case the dichotomic measurement $A'$ is further decomposed into a measurement having three distinct outcomes, transforming the implicit alternative into an explicit one. More precisely, this fully unpacked measurement, which we denote by $A''$, now has the three outcomes $A''_{yy}$, $A''_{yn}$ and $A''_n$, and the  associated states $p_{A''_{yy}}$, $p_{A''_{yn}}$ and $p_{A''_n}$, respectively,  where $A''_n=A_n$, $A''_{yy}=A_y\wedge B_y$ and $A''_{yn}= A_y\wedge B_n$. Thus, participants can choose among three distinct possibilities, with probabilities $p_S(A''_i)=\mu(p_{A''_i}, e_{A''},p_S)$, $i\in\{yy,yn,n\}$. Again, one speaks of superadditivity if $p_S(A_y)>p_S(A''_{yy})+p_S(A''_{yn})$ and of subadditivity if $p_S(A_y)<p_S(A''_{yy})+p_S(A''_{yn})$.

Since superadditivity and subadditivity are in general both possible, the usual quantum analysis exploits the interference effects as a way to explain, by means of a single mechanism, both possibilities, as interference terms can take both positive and negative values \cite{Franco2008}. The assumption behind this approach is that participants act in a sequential way, all with the same order for the sequence. Accordingly, one associates the non-commuting projection operators $P^A_i$ and $P^B_j$ to the outcomes $A_i$ and $B_j$, respectively, $i,j\in\{y,n\}$, so that one can write, for every $i\in\{y,n\}$,
\begin{equation}
P^A_i = P^B_yP^A_iP^B_y + P^B_nP^A_iP^B_n + I_i  
\label{sequential}
\end{equation}
where $P^B_n = \mathbb{I}-P^B_y$ and $I_i=P^B_yP^A_iP^B_n + P^B_n P^A_iP^B_y$ is the interference contribution, responsible of the superadditivity or subadditivity effects. 

The above analysis, however, has some weak points. Firstly, the above projection operators do not commute, hence the order of evaluation in the sequence becomes important, and one needs to assume that all participants always start by answering first the question $B$ and only then the question $A$. However, since this sequentiality is not part of the experimental protocol, nothing guarantees that it will be carried out in practice, instead of considering $A''_{yy}$ and $A''_{yn}$ as outcomes of a single non-sequential measurement. Secondly, it is incompatible with the natural interpretation of the packed and explicitly unpacked  outcomes as belonging to  two measurements that are logically related, in the sense that $A$ can be understood as the degenerate version of the non-degenerate measurement $A''$ or, to put it another way, as a sub-measurement of $A''$. 

Considering  the packed measurement $A$ and the associated explicitly unpacked measurement $A''$, the question is: How should we use the quantum formalism to model these experimental situations? In both measurements we have a cognitive entity in the same initial state $p_S$. We also have outcomes that are the same for both measurements, $A_n$ and $A''_n$, which therefore should be associated with the same state, describing the same intersubjective reality. Then, we have outcomes that are described in a packed way in one measurement and in an explicitly unpacked way in the other -- in our example the outcome $A_y$ that is decomposed into the two alternatives $A''_{yy}$ and $A''_{yn}$. 

If quantum theory is taken as a unitary and coherent framework, one should then be able to use the notion of `degenerate measurement' (the quantum notion of sub-measurement) to model these two logically related experimental situations. Considering the previous example of the  entity {\it The winner of next US presidential election}, it is clear that a `non-Democrat' president is either a `Republican' or an `Independent,' and that `Republican' and `Independent' presidents are always `non-Democrat' presidents. This means that the `Republican' or the `Independent' specification is an additional specification for the `non-Democrat' state, and this means that when comparing an experimental situation where this specification is made, to a situation where it is not made, the latter should be considered as a sub-measurement of the former, i.e. a degenerate measurement in the quantum jargon. Indeed, when the outcome is just `Non-Democrat', the experimenter has no information about the `Independent' or `Republican' element, this being not specified in the outcome state. Also, since `Republican' and `Independent' are excluding possibilities, within the quantum formalism one should certainly describe them by two orthogonal subspaces, or two orthogonal states. Considering all this, one would thus expect to get
\begin{equation}
p_S(A_y)=p_S(A''_{yy})+p_S(A''_{yn}) \qquad p_S(A_n)= p_S(A''_n).
\label{equality}
\end{equation}
But, since explicit unpacking effects are observed (which are generally stronger than the implicit ones), equalities like the above can be expected to be  significantly violated, meaning that sub-measurements in psychology would not allow themselves to be consistently represented in the Hilbert space quantum formalism. Again, this can be attributed to the fact that the latter only admits a single `way of choosing' the available outcomes, the `Born way', whereas it is more natural to assume that the selection process can generally depend on the overall cognitive situation that is presented to the participants. Indeed, participants' propensity of choosing a given outcome certainly depends on the nature of the alternatives that are presented to them, and this is a contextuality effect that the quantum formalism is unable to describe. Yet, it can be represented in the GTR-model and its EBR implementation, by assuming that the probability distribution $\rho_{A}$ characterizing the degenerate measurement $A$ is not  the same as the probability distribution $\rho_{A''}$ describing the corresponding non-degenerate versions $A''$, associated with an explicitly unpacked situation. 

Concerning the implicitly unpacked case, one would also expect, if the standard quantum formalism applied, that $p_S(A_n)= p_S(A'_n)$, implying that $p_S(A_y)=p_S(A'_y)$. However, since the packed and implicitly unpacked measurements are dichotomic measurements, sharing the same state $p_{A_n} =p_{ A'_n}$, if follows that the two states corresponding to the outcomes $A_y$ and $A'_y=(A_y\wedge B_y)\oplus (A_y\wedge B_n)$ should also be equal, implying the equality of the associated transition probabilities. Thus, also in this case a Hilbert space quantum formalism cannot be used to model the data. In fact, even the EBR is too specific in this case, as it also relies on the Hilbert space structure for the representation of states (in the EBR, if two non-degenerate two-outcome measurements share an eigenstate, they necessarily also share the other one, as to each point on the three-dimensional Bloch sphere there is only one corresponding antipodal point). This is a situation where the more general GTR-model is required \cite{asdb2015f},  as it allows one to 
describe two dichotomic measurements by means of two probability distributions defined on line segments that share one of their vertex points (corresponding to the outcome $A_n$), but not the other.

To conclude, we observe that in \cite{RottenstreichTversky1997} the protocol was such that respondents were partitioned in four groups, each group responding one of the four different `yes/no' alternatives for {\it The winner of the next US presidential election}: `Non-Democrat', `Independent rather than Republican or Democrat', `Republican rather than Independent or Democrat' and `Independent or Republican rather than Democrat'. 
In other terms, they were actually performing a single measurement with five distinct outcomes. The experimental situation we have discussed is different, although of course related, and to apply the data to our analysis one should repeat the experiment by splitting it into three measurements: ($A$) one with outcomes `non-Democrat' and `Democrat'; ($A'$) another one with outcomes `Independent or Republican rather than a Democrat' and `Democrat'; ($A''$) and a last one with outcomes `Independent', `Republican' and `Democrat'.

\end{document}